\documentclass{article}

\usepackage[nonatbib,final]{nips_2017}

\usepackage[utf8]{inputenc} 
\usepackage[T1]{fontenc}    
\usepackage{hyperref}       
\usepackage{url}            
\usepackage{booktabs}       
\usepackage{amsfonts}       
\usepackage{nicefrac}       
\usepackage{microtype}      

\usepackage{graphicx}
\usepackage{lipsum}
\usepackage{amsmath}
\usepackage{caption}
\usepackage{subcaption}

\title{fpgaConvNet: A Toolflow for Mapping Diverse Convolutional Neural Networks on Embedded FPGAs}

\author{
  Stylianos I. Venieris 
  \\
  Dept. of Electrical and Electronic Engineering\\
  Imperial College London\\
  \texttt{stylianos.venieris10@ic.ac.uk} \\
   \And
   Christos-Savvas Bouganis \\
   Dept. of Electrical and Electronic Engineering \\
   Imperial College London \\
   \texttt{christos-savvas.bouganis@ic.ac.uk} \\
}

\begin{document}

\maketitle
\begin{abstract}
In recent years, Convolutional Neural Networks (ConvNets) have become an enabling technology for a wide range of novel embedded Artificial Intelligence systems. Across the range of applications, the performance needs vary significantly, from high-throughput video surveillance to the very low-latency requirements of autonomous cars. In this context, FPGAs can provide a potential platform that can be optimally configured based on the different performance needs. However, the complexity of ConvNet models keeps increasing making their mapping to an FPGA device a challenging task. This work presents fpgaConvNet, an end-to-end framework for mapping ConvNets on FPGAs. The proposed framework employs an automated design methodology based on the Synchronous Dataflow (SDF) paradigm and defines a set of SDF transformations in order to efficiently explore the architectural design space. By selectively optimising for throughput, latency or multiobjective criteria, the presented tool is able to efficiently explore the design space and generate hardware designs from high-level ConvNet specifications, explicitly optimised for the performance metric of interest. Overall, our framework yields designs that improve the performance by up to 6.65$\times$ over highly optimised embedded GPU designs for the same power constraints in embedded environments.

%
\end{abstract}


\section{Introduction}
\label{intro}
In recent years, the Deep Learning model of Convolutional Neural Networks (ConvNets) has pushed the boundaries of several Artificial Intelligence tasks. From object tracking \cite{Held2016} to drone trail navigation \cite{Smolyanskiy_2017}
, ConvNets have been an enabling technology behind a wide variety of applications. In the embedded space, novel complex systems such as autonomous drones and self-driving cars are employing multiple ConvNets in order to perceive their surroundings and ultimately execute their high-level tasks. At the same time, embedded systems pose stringent requirements with respect to throughput, latency and power consumption which becomes a challenge in the case of computationally heavy ConvNets. With general purpose parallel architectures reaching the limit of satisfying these constraints, specialised hardware solutions are becoming a necessity. 

In this context, reconfigurable hardware in the form of Field-Programmable Gate Arrays (FPGAs) emerges as a promising alternative. FPGAs offer the benefits of customisability and reconfigurability by means of a set of heterogeneous hardware resources with programmable interconnections between them. 
With FPGAs' size and resource specifications advancing at a fast pace and with ConvNets becoming more complex, the possible mappings of a ConvNet on an FPGA lie in a large multidimensional design space that cannot be explored manually. At the same time, the diversity of ConvNet application domains results in a wide spectrum of performance needs. 
To this end, there is a need for tools that abstract the low-level resource details of a particular FPGA and automate the mapping of ConvNets on FPGAs in a principled manner while satisfying the application-level performance needs. 


In this context, we present fpgaConvNet \cite{Venieris_2016}, an automated ConvNet-to-FPGA framework that bridges the gap between the existing Deep Learning software ecosystem and FPGAs. The presented framework consists of the following features:

\begin{itemize}
\item Supporting a Caffe front end and capturing analytically by means of a Synchronous Dataflow (SDF) model the ConvNet workload and the target FPGA resource budget. 

\item Formulating the design space exploration as a mathematical optimisation problem and optimising with respect to throughput, latency or multiobjective criteria based on the application-level performance requirements.

\item Generating optimised streaming accelerators for ConvNet models with both regular layer connectivity, such as AlexNet \cite{Krizhevsky2012} 
and VGG16 \cite{Simonyan14c}, and models with irregular dataflow, including inception-based \cite{Szegedy2014}, residual \cite{He_2016} and dense \cite{huang2017densely} networks. Moreover, we present the first mapping of DenseNet-161 on FPGAs.

\item Demonstrating 
up to 6.65$\times$ higher performance over highly optimised GPU designs for the same power constraints in embedded environments.

\end{itemize}

\begin{figure}[b]
	\vspace{-0.65cm}
	\centering
	\begin{subfigure}{.45\linewidth}
		\centering
		{\includegraphics[trim={0cm 12cm 0cm 13cm},clip,width=1\linewidth]{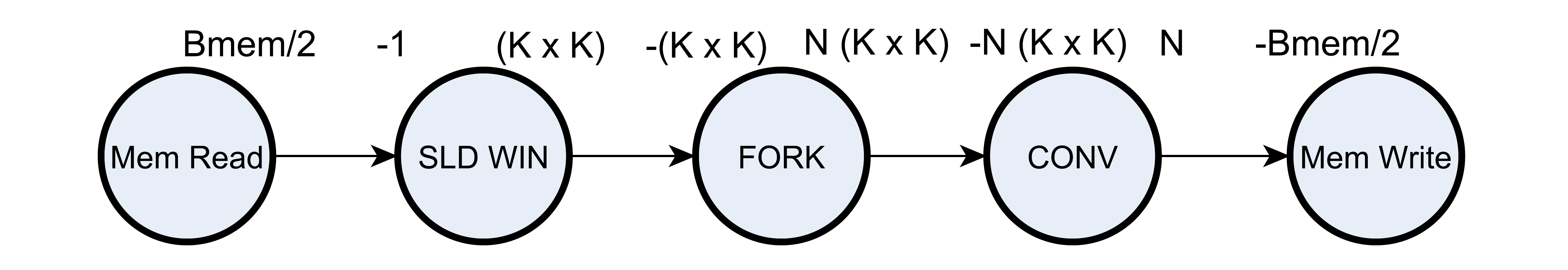}}
		\begin{equation*}
		\resizebox{1\linewidth}{!}{%
			$\boldsymbol{\Gamma} = 
			\left[\begin{matrix}
			\frac{B_{mem}}{2} & -1 & 0 & 0 & \phantom{-}0 \\ 
			0 & (K\times K) & -(K\times K) & 0 & \phantom{-}0 \\
			0 & 0 & N (K \times K) & -N (K \times K) & \phantom{-}0 \\
			0 & 0 & 0 & N & -\frac{B_{mem}}{2} \\
			\end{matrix}\right]$
		}
		\label{topology_matrix_example_1}
		\end{equation*}
		\vspace{-0.25cm}	
		\caption{ }
		\label{sdf_graph_example_1}
	\end{subfigure}
	\begin{subfigure}{0.45\linewidth}
		
		
		\centering
		{\includegraphics[trim={0cm 7cm 0cm 9.25cm},clip,width=1\linewidth]{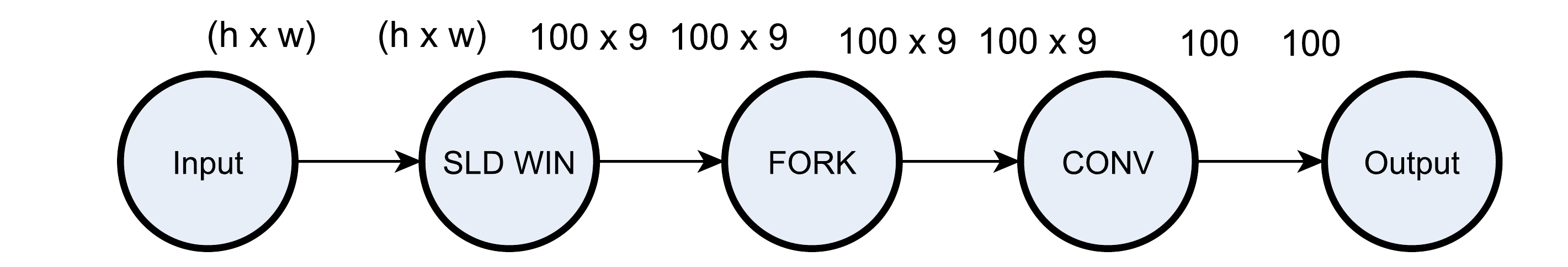}}
		
		\vspace{-0.35cm}
		\begin{equation*}
		\resizebox{0.85\linewidth}{!}{
			$\boldsymbol{W} = 
			\left[\begin{matrix}
			h \times w & h \times w & 0 & 0 & 0 \\
			0 & 100 \times 9 & 100 \times 9 & 0 & 0 \\ 
			0 & 0 & 100 \times 9 & 100 \times 9 & 0 \\
			0 & 0 & 0 & 100 & 100 \\
			\end{matrix}\right]$
		}
		\label{dot_product_example_2_1}
		\end{equation*}
		\vspace{-0.3cm}
		\caption{ } 
		\label{sdf_workload_example_2}
	\end{subfigure}
	\caption{Illustration of representing the hardware mapping (\ref{sdf_graph_example_1}) and the workload (\ref{sdf_workload_example_2}) of a convolutional layer by means of fpgaConvNet's SDF model}
\end{figure}

\section{Synchronous Dataflow Modelling for ConvNets}
\subsection{Modelling ConvNets with SDF}
By interpreting ConvNet inference as a streaming application, fpgaConvNet employs the Synchronous Dataflow model of computation as its modelling core. Synchronous Dataflow (SDF) \cite{Lee_1987} is widely used for the analysis and design of parallel systems. Under this scheme, a computing system is modelled as a directed graph (SDFG), with the nodes representing computations and with arcs in place of data streams between them. The basic principle of SDF is the data-driven execution where each node fires whenever data are available at its incoming arcs. In this context, we propose and use an SDF model for capturing ConvNet workloads and representing hardware mappings by means of linear algebra and graph theory. This formulation enables us to capture each design point with a number of compile-time configurable parameters and explore efficiently the architectural design space by means of a set of proposed transformations. The transformations can be expressed as algebraic operations and applied directly over the SDF model of a design point in order to modify the performance-resource characteristics of the potential implementation. Moreover, the proposed modelling approach enables us to formally express the design space exploration as a mathematical optimisation problem.

\subsection{ConvNet Hardware Mappings as SDF Graphs}
At a hardware level, fpgaConvNet represents design points as SDF graphs that can execute the input ConvNet workload. Given a target ConvNet, each layer is mapped to a sequence of \textit{hardware building blocks} that implement the layer's functionality. By assigning one SDF node to each building block, an SDF graph is constructed. The nodes of the SDFG are connected via arcs which carry data between building blocks. Each  building block is defined by a set of parameters that can be configured at compile time. This process leads to the formation of a hardware architecture which consists of a coarse pipeline of building blocks and corresponds to a design point in the architectural design space. The SDF graph of a design point can be represented equivalently by means of a topology matrix $\boldsymbol{\Gamma}$$\in$$\mathbb{R}^{(M \times N)}$ parametrised over the tunable parameters of each instantiated building block, where $M$ and $N$ are the number of nodes and arcs in the SDFG respectively. The element $\boldsymbol{\Gamma}$($a$,$n$) holds the data rate of node $n$ at arc $a$, with positive and negative sign for data production and consumption respectively. Fig. \ref{sdf_graph_example_1} shows an example of the graph and matrix representations of a convolutional layer's hardware mapping.

\subsection{ConvNet Workloads as SDF Graphs}
A ConvNet workload is represented as a stream of data flowing through a sequence of building blocks. 
By creating a \textit{workload matrix} $\boldsymbol{W}$ whose columns hold the local workload at the input and output of each building block in the architecture, a compact and distributed representation of the computational workload can be constructed (Fig. \ref{sdf_workload_example_2}). The amount of work, $W^{in}_i$ and $W^{out}_i$, carried out by the $i_{th}$ hardware block during ConvNet inference is defined as the total number of data elements to be consumed and produced by this block respectively.

\subsection{SDF Transformations for Efficient Design Space Exploration}
\begin{figure}[t]
	\vspace{-1cm}
	\centering
	{\includegraphics[trim={0cm 10.25cm 0cm
	1.25cm},clip,width=1\linewidth]{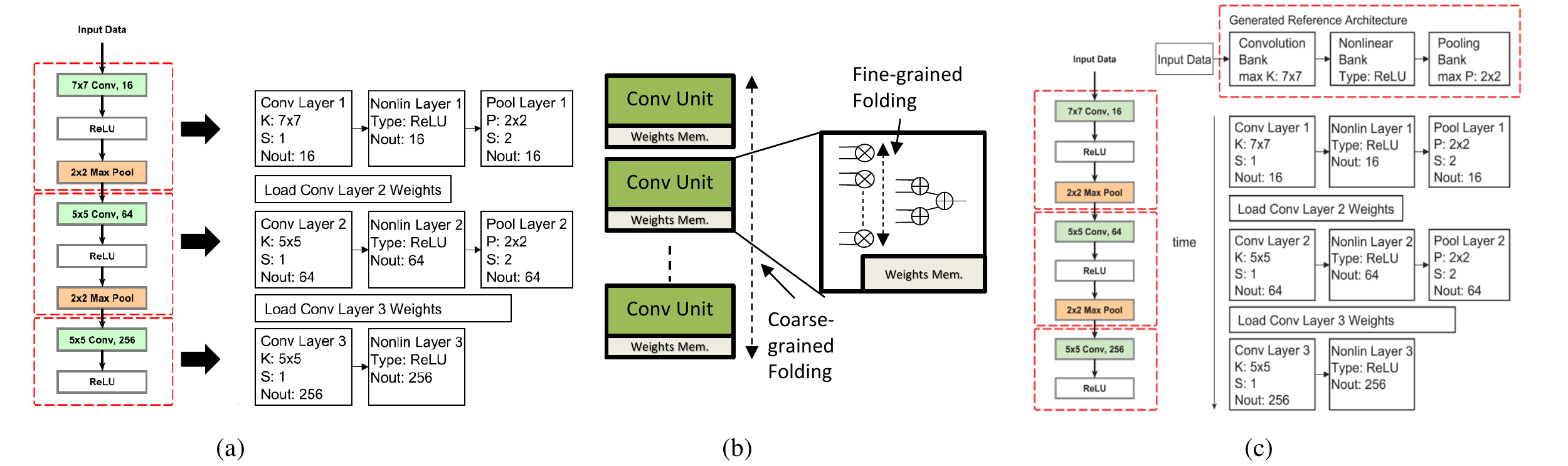}}
	\vspace{-0.6cm}
	\caption{SDF Transformations: (a) graph partitioning with reconfiguration, (b) coarse- and fine-grained folding, (c) weights reloading.}
	\label{sdf_transformations}
	\vspace{-0.65cm}
\end{figure}

To modify the configurable parameters of each hardware building block, fpgaConvNet introduces four transformations: (1) \textit{graph partitioning with reconfiguration}, (2) \textit{coarse-grained folding}, \mbox{(3) \textit{fine-grained folding}} and (4) \textit{weights reloading}. Graph partitioning with reconfiguration is tailored to high-throughput applications and achieves high throughput by partitioning a ConvNet along its depth and constructing one SDF subgraph per partition. One distinct architecture is generated per subgraph tailored to the subgraph's workload, which can exploit all the resources of the target FPGA to reach high performance. The execution of each subgraph requires the full reconfiguration of the FPGA and fpgaConvNet amortises the reconfiguration time overhead by means of batch processing, leading to high-throughput mappings (Fig. \ref{sdf_transformations}a).
Coarse-grained folding enables the tuning of the number of coarse units for each layer, spanning from a fully parallel implementation down to a single, time-shared compute unit (Fig. \ref{sdf_transformations}b). Fine-grained folding allows the configuration of the dot product implementation inside the convolutional units of a layer, spanning from a fully parallel implementation down to a single, time-shared multiply-accumulate  unit (Fig. \ref{sdf_transformations}b). The weights reloading transformation is tailored to latency-sensitive applications and provides a method of executing several subgraphs without adding a penalty on latency due to full FPGA reconfiguration \cite{Venieris_2017b}. Similarly to graph partitioning with reconfiguration, this transformation partitions the SDFG into several subgraphs. However, instead of generating a distinct, fixed
architecture for each subgraph, a single flexible \textit{reference} architecture is derived, by holistically optimising across all the subgraphs, which is capable of executing the workloads of all the subgraphs by operating in different modes \mbox{(Fig. \ref{sdf_transformations}c)}. 

We express all transformations as algebraic operations that can be applied directly over the SDF model. In this manner, we build analytical throughput and latency estimators for design points together with resource consumption models parametrised with respect to the tunable parameters of each building block, and cast the design space exploration to an optimisation problem.

\section{Evaluation}

\subsection{Comparison with Embedded GPUs}

In power-constrained mobile and embedded ConvNet applications, the primary metrics of interest comprise (1) the absolute power consumption and (2) the performance efficiency in terms of performance-per-Watt. In this respect, we investigate the performance efficiency of fpgaConvNet by targeting the Xilinx Zynq 7045 FPGA which is an industry standard for mobile applications and comparing with the widely used, high-performance low-power NVIDIA Tegra X1 platform. 

Our evaluation is focused on ConvNets that are commonly used as pre-trained models and includes widely employed networks with regular layer connectivity (AlexNet, VGG16) and irregular dataflow (GoogLeNet, ResNet-152, DenseNet-161). For the execution of the benchmarks on Tegra X1, we use the TensorRT framework with cuDNN v6 and FP16 precision in order to obtain highly optimised mappings. For the FPGA, we employ 16-bit fixed-point arithmetic (FXP16). Furthermore, power measurements for both platforms 
are obtained via a power monitor 
and we subtract the 
idle power in order to obtain the power due to benchmark execution.


\begin{figure}[t]
	\vspace{-0.65cm}
	\begin{subfigure}{0.5\linewidth}
		\centering
		{\includegraphics[trim={7cm 16.75cm 7cm 16.5cm},clip,width=1\linewidth]{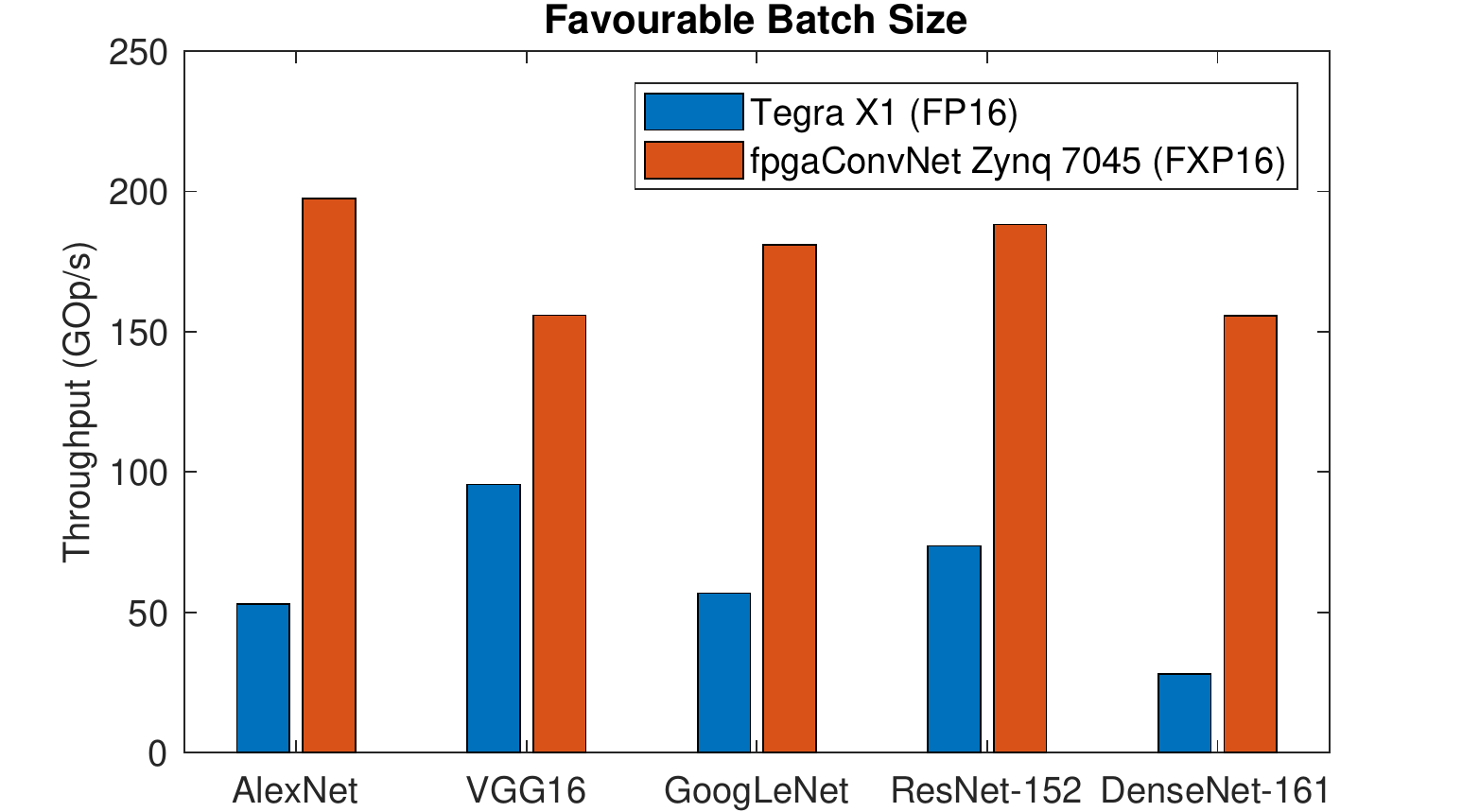}}
		\caption{Evaluation on high-throughput applications with favourable batch size.}
		\label{fpga_vs_tx1_high_throughout}
		\vspace{-0.15cm}
	\end{subfigure}
	\hspace{0.1cm}
	\begin{subfigure}{0.5\linewidth}
		\centering
		{\includegraphics[trim={7cm 16.75cm 7cm 16.5cm},clip,width=1\linewidth]{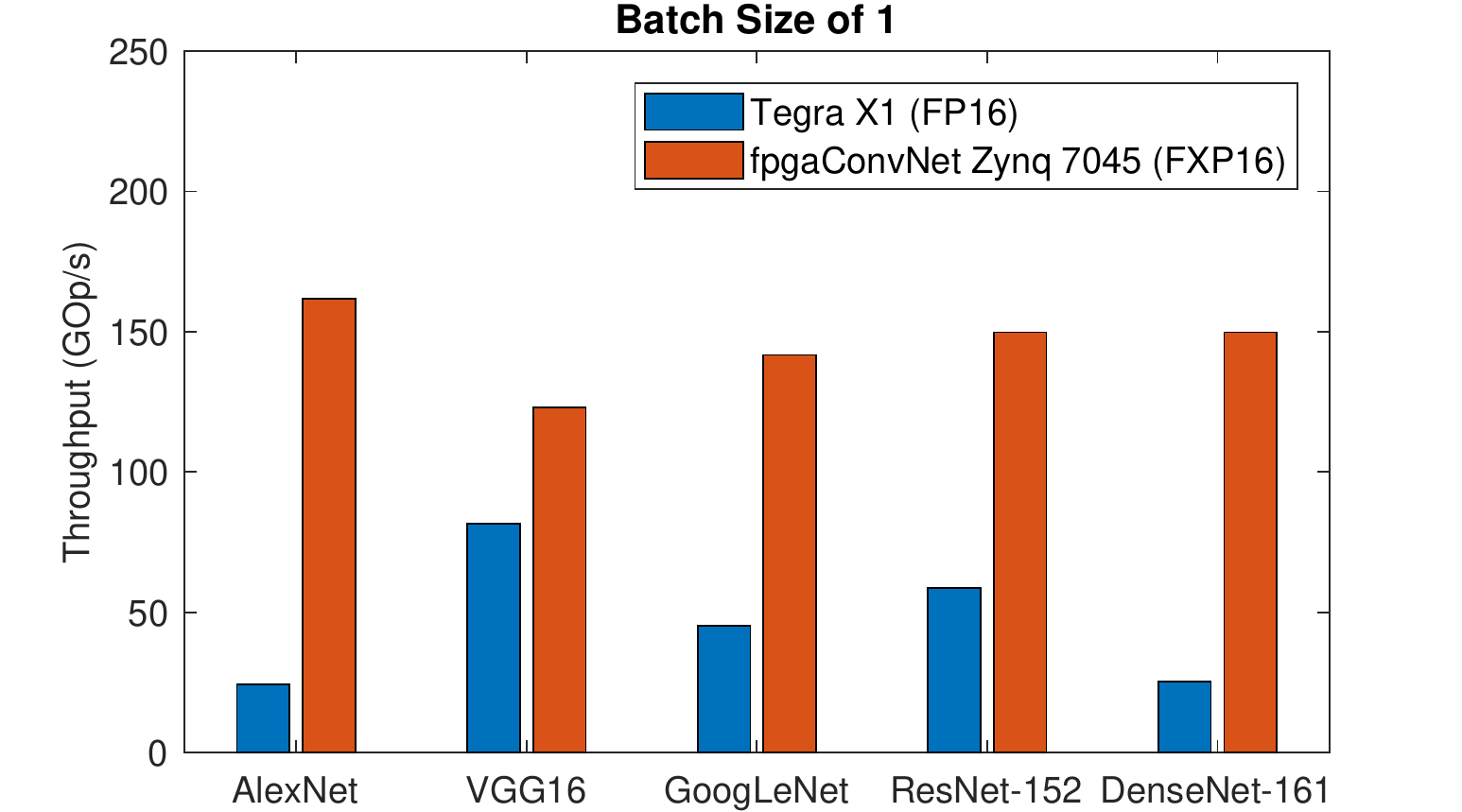}}
		\caption{Evaluation on latency-sensitive applications with batch size of 1.}
		\label{fpga_vs_tx1_low_latency}
		\vspace{-0.15cm}
	\end{subfigure}
	\caption{Comparison of fpgaConvNet on Zynq 7045 with Tegra X1 on throughput-driven (a) and latency-driven (b) applications. Both platforms are evaluated with a power budget of 5 W.}
	\vspace{-0.5cm}
\end{figure}

%

\vspace{-0.25cm}

\paragraph{Discussion.}
Tegra X1 mounts a 256-core GPU and supports a range of clock frequencies up to \mbox{998 MHz,} with a peak of 1024 GFLOPS at around 15 W with FP16. To investigate the performance of each platform under the same absolute power constraints that would be present in an embedded setting, we configure the frequency of the GPU with 76.8 MHz and the FPGA at 125 MHz for the same power budget of 5 W. With this setup, Tegra X1 and Zynq 7045 have a peak performance of 79 GFLOPS and 270 GOp/s with a memory bandwidth of 12.8 GB/s and 4.2 GB/s respectively. Fig. \ref{fpga_vs_tx1_high_throughout} and \ref{fpga_vs_tx1_low_latency} show the measured throughput for each ConvNet on Tegra X1 and Zynq 7045 for throughput-driven and latency-driven applications respectively. For throughput-driven applications, fpgaConvNet achieves a throughput improvement over Tegra X1 of up to 5.53$\times$ with an average of 3.32$\times$ (3.07$\times$ geo. mean) across the benchmarks. For latency-driven scenarios, fpgaConvNet demonstrates a throughput improvement of up to 6.65$\times$ with an average of 3.95$\times$ \mbox{(3.43$\times$ geo. mean).} To investigate the peak performance-per-Watt, we run the same benchmarks on the GPU with maximum frequency of 998 MHz. In this setting, for throughput-driven applications, fpgaConvNet achieves 1.17$\times$\mbox{(1.12$\times$ geo. mean)} improvement in GOp/s/W and for latency-driven applications, fpgaConvNet achieves 1.70$\times$ (1.36$\times$ geo. mean) over Tegra X1. Based on the presented evaluation, fpgaConvNet demonstrates gains in performance-per-Watt across the benchmarks and reaches higher raw performance over highly optimised GPU mappings when comparable power constraints are present as is common in mobile and embedded systems.

\section{Conclusion}
The deployment of Convolutional Neural Networks in the embedded space is posing challenges due to the computational complexity of ConvNets and the stringent performance and power constraints of mobile and embedded applications. In this respect, the development of specialised hardware solutions for the efficient processing of ConvNets has emerged as a promising approach. In this context, fpgaConvNet is a framework that automates the optimised mapping of Convolutional Neural Networks on FPGAs, targeting models with both regular and irregular layer connectivity. 
fpgaConvNet introduces an SDF-based design methodology in order to traverse efficiently the FPGA architectural design space. By casting the design space exploration task as a multiobjective optimisation problem, fpgaConvNet is able to effectively target applications with a variety of performance needs, from high throughput to low latency. 
Quantitative evaluation demonstrates that the fpgaConvNet-generated accelerators manage to achieve higher performance-per-Watt than highly optimised embedded GPU designs and reach higher raw performance under the same power constraints, and therefore provides the infrastructure for the deployment of ConvNet models on embedded FPGAs.

\small

\bibliographystyle{IEEEtran}
\bibliography{Bibliography}

\begin{thebibliography}{10}
\providecommand{\url}[1]{#1}
\csname url@samestyle\endcsname
\providecommand{\newblock}{\relax}
\providecommand{\bibinfo}[2]{#2}
\providecommand{\BIBentrySTDinterwordspacing}{\spaceskip=0pt\relax}
\providecommand{\BIBentryALTinterwordstretchfactor}{4}
\providecommand{\BIBentryALTinterwordspacing}{\spaceskip=\fontdimen2\font plus
\BIBentryALTinterwordstretchfactor\fontdimen3\font minus
  \fontdimen4\font\relax}
\providecommand{\BIBforeignlanguage}[2]{{%
\expandafter\ifx\csname l@#1\endcsname\relax
\typeout{** WARNING: IEEEtran.bst: No hyphenation pattern has been}%
\typeout{** loaded for the language `#1'. Using the pattern for}%
\typeout{** the default language instead.}%
\else
\language=\csname l@#1\endcsname
\fi
#2}}
\providecommand{\BIBdecl}{\relax}
\BIBdecl

\bibitem{Held2016}
D.~Held, S.~Thrun, and S.~Savarese, ``{Learning to Track at 100 FPS with Deep
  Regression Networks},'' in \emph{ECCV}, 2016.

\bibitem{Smolyanskiy_2017}
N.~Smolyanskiy, A.~Kamenev, J.~Smith, and S.~Birchfield, ``{Toward Low-Flying
  Autonomous {MAV} Trail Navigation using Deep Neural Networks for
  Environmental Awareness},'' in \emph{IROS}, 2017.

\bibitem{Venieris_2016}
S.~I. Venieris and C.-S. Bouganis, ``{fpgaConvNet: A Framework for Mapping
  Convolutional Neural Networks on FPGAs},'' in \emph{{FCCM}}, 2016.

\bibitem{Krizhevsky2012}
A.~Krizhevsky, I.~Sutskever, and G.~E. Hinton, ``{ImageNet Classification with
  Deep Convolutional Neural Networks},'' in \emph{NIPS}, 2012.

\bibitem{Simonyan14c}
K.~Simonyan and A.~Zisserman, ``{Very Deep Convolutional Networks for
  Large-Scale Image Recognition},'' in \emph{ICLR}, 2015.

\bibitem{Szegedy2014}
C.~Szegedy \emph{et~al.}, ``{Going Deeper with Convolutions},'' in \emph{CVPR},
  2015.

\bibitem{He_2016}
K.~He, X.~Zhang, S.~Ren, and J.~Sun, ``{Deep Residual Learning for Image
  Recognition},'' in \emph{{CVPR}}, 2016.

\bibitem{huang2017densely}
G.~Huang, Z.~Liu, L.~van~der Maaten, and K.~Q. Weinberger, ``{Densely Connected
  Convolutional Networks},'' in \emph{CVPR}, 2017.

\bibitem{Lee_1987}
E.~A. Lee and D.~G. Messerschmitt, ``{Synchronous Data Flow},''
  \emph{Proceedings of the IEEE}, Sept 1987.

\bibitem{Venieris_2017b}
S.~I. Venieris and C.-S. Bouganis, ``{Latency-Driven Design for FPGA-based
  Convolutional Neural Networks},'' in \emph{FPL}, 2017.

\bibitem{Paszke2016}
A.~Paszke, A.~Chaurasia, S.~Kim, and E.~Culurciello, ``{ENet: {A} Deep Neural
  Network Architecture for Real-Time Semantic Segmentation},'' \emph{CoRR},
  2016.

\bibitem{Cavigelli2017}
L.~Cavigelli, P.~Degen, and L.~Benini, ``{CBinfer: Change-Based Inference for
  Convolutional Neural Networks on Video Data},'' in \emph{ICDSC}, 2017.

\end{thebibliography}

\end{document}